\documentclass[10pt,twocolumn,letterpaper]{article}

\usepackage{cvpr}              % To produce the CAMERA-READY version
\usepackage{graphicx}
\usepackage{amsmath}
\usepackage{amssymb}
\usepackage{booktabs}
\usepackage{pifont}% http://ctan.org/pkg/pifont
\newcommand{\cmark}{\ding{51}}%
\newcommand{\xmark}{\ding{55}}%
\usepackage{makecell}
\usepackage{colortbl}
\definecolor{mygray}{gray}{.9}
\definecolor{mypink}{rgb}{.99,.91,.95}
\definecolor{mycyan}{cmyk}{.3,0,0,0} 

\usepackage{xcolor}
\definecolor{citecolor}{HTML}{0071bc} 
\definecolor{SeaGreen4}{RGB}{0,205,102} 
\definecolor{SlateBlue}{RGB}{106,90,205} 
\definecolor{DarkRed}{RGB}{178,34,34}

\usepackage{threeparttable}
\usepackage{multirow}
\usepackage{longtable}
\usepackage{rotating}
\usepackage{tabularx}
\usepackage{color}
\usepackage{xcolor}
\definecolor{citecolor}{HTML}{0071bc}
\usepackage[colorlinks, linkcolor=red,  anchorcolor=blue, citecolor=citecolor]{hyperref}

\usepackage[capitalize]{cleveref}
\crefname{section}{Sec.}{Secs.}
\Crefname{section}{Section}{Sections}
\Crefname{table}{Table}{Tables}
\crefname{table}{Tab.}{Tabs.}

\begin{document}

\title{ HARDVS: Revisiting Human Activity Recognition with Dynamic Vision Sensors } 

\author{Xiao Wang$^{1}$, Zongzhen Wu$^{1}$, Bo Jiang$^{1}$\thanks{Corresponding author: Bo Jiang}, Zhimin Bao$^{2}$, Lin Zhu$^{3}$ \\ 
Guoqi Li$^{4}$, Yaowei Wang$^{5}$, Yonghong Tian$^{5,6}$ \\ 
${^1}$\emph{School of Computer Science and Technology, Anhui University, Hefei, China} \\
${^2}$\emph{Tencent YouTu Lab, Hefei, China} ~~~
${^3}$\emph{Beijing Institute of Technology, Beijing, China} \\
${^4}$\emph{Institute of Automation, Chinese Academy of Sciences, Beijing, China} \\ 
${^5}$\emph{Peng Cheng Laboratory, Shenzhen, China} ~~~
${^6}$\emph{Peking University, Beijing, China} 
}

\maketitle

%%%%%%%%% ABSTRACT
\begin{abstract}
The main streams of human activity recognition (HAR) algorithms are developed based on RGB cameras which are suffered from illumination, fast motion, privacy-preserving, and large energy consumption. Meanwhile, the biologically inspired event cameras attracted great interest due to their unique features, such as high dynamic range, dense  temporal but sparse spatial resolution, low latency, low power, etc. As it is a newly arising sensor, even there is no realistic large-scale dataset for HAR. Considering its great practical value, in this paper, we propose a large-scale benchmark dataset to bridge this gap, termed HARDVS, which contains 300 categories and more than 100K event sequences. 
% HARDVS fully reflects the features of event cameras by considering factors including multiple views, illumination, motion speed, dynamic background, occlusion, glitter, and photographic distance. 
We evaluate and report the performance of multiple popular HAR algorithms, which provide extensive baselines for future works to compare. 
More importantly, we propose a novel spatial-temporal feature learning and fusion framework, termed ESTF, for event stream based human activity recognition. It first projects the event streams into spatial and temporal embeddings using StemNet, then, encodes and fuses the dual-view representations using Transformer networks. 
Finally, the dual features are concatenated and fed into a classification head for activity prediction. Extensive experiments on multiple datasets fully validated the effectiveness of our model. 
Both the dataset and source code will be released on \url{https://github.com/Event-AHU/HARDVS}. 
\end{abstract}

%%%%%%%%% BODY TEXT
\section{Introduction} 

With the rapid development of the smart city, recognizing human behavior (i.e., Human Activity Recognition, HAR) accurately and efficiently is becoming an extremely urgent task. Most researchers develop the HAR algorithms~\cite{kong2018humanARSurvey, ahmad2021graph} based on RGB cameras which are widely deployed and easy to collect the data. With the help of large-scale benchmark datasets~\cite{gu2018ava, kay2017kinetics, caba2015activitynet, kuehne2011hmdb, monfort2019moments, soomro2012ucf101, sigurdsson2016hollywood} and deep learning, HAR in regular scenarios has been studied to some extent. However, the storage, transmission, and analysis of surveillance videos set limits the demands for the practical systems due to the usage of RGB sensors. More in detail, the standard RGB cameras have a limited frame rate (e.g., 30 FPS) which makes it hard to capture the fast-moving objects and is easily influenced by motion blur. The low dynamic range (60 dB) makes the RGB sensors work poorly in low illumination. It also suffers from the high redundancy between nearby frames which needs more storage and energy   consumption. 
Privacy protection also greatly limits its development, therefore, a natural question is \emph{do we have to recognize human activities using the RGB sensors?}

\begin{figure} 
\center
\includegraphics[width=3.3in]{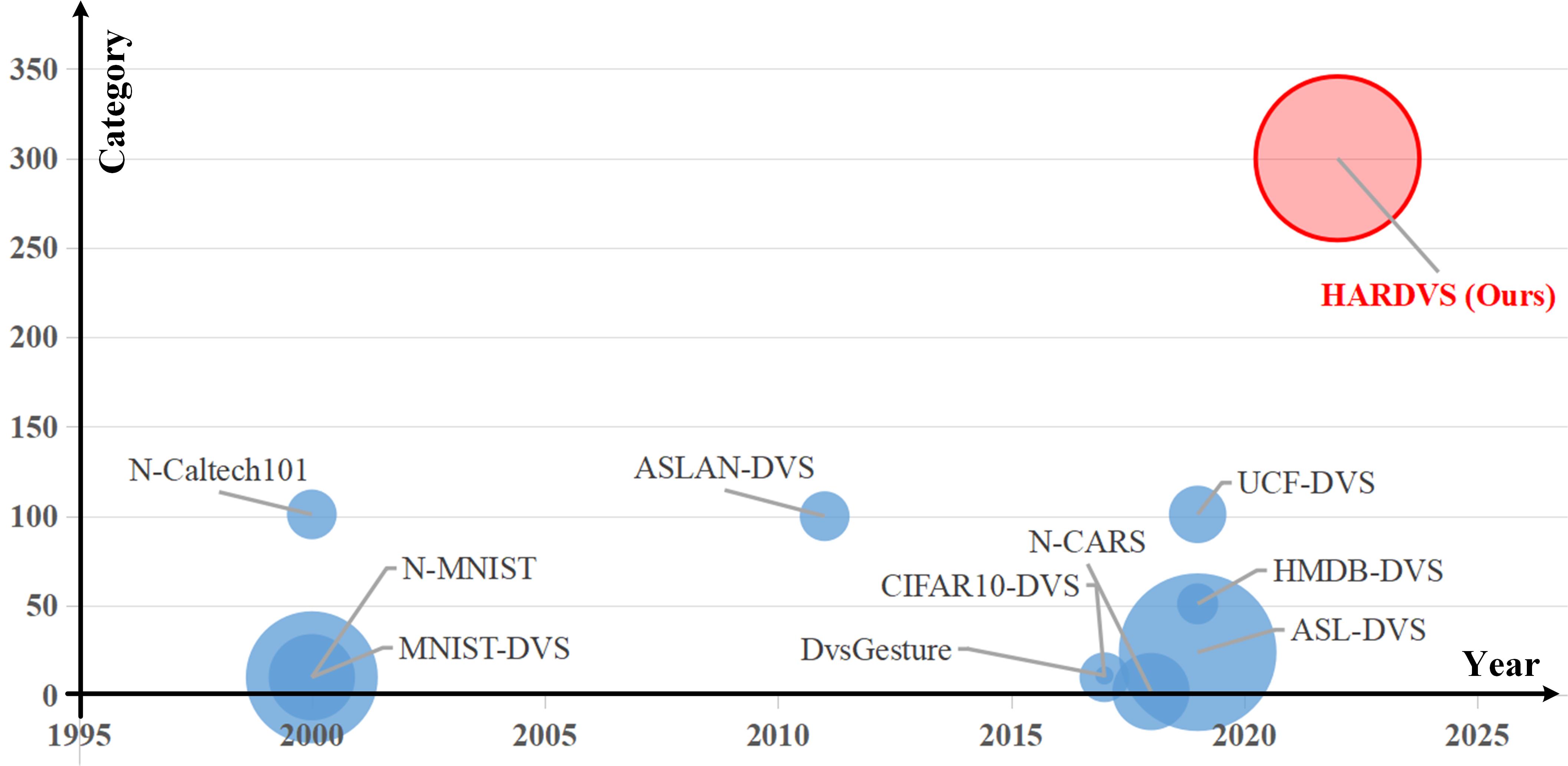}
\caption{Comparison between existing datasets and our proposed HARDVS dataset for event based video classification.} 
\label{benchmarkCOMP}
\end{figure}

\begin{figure} 
\center
\includegraphics[width=3in]{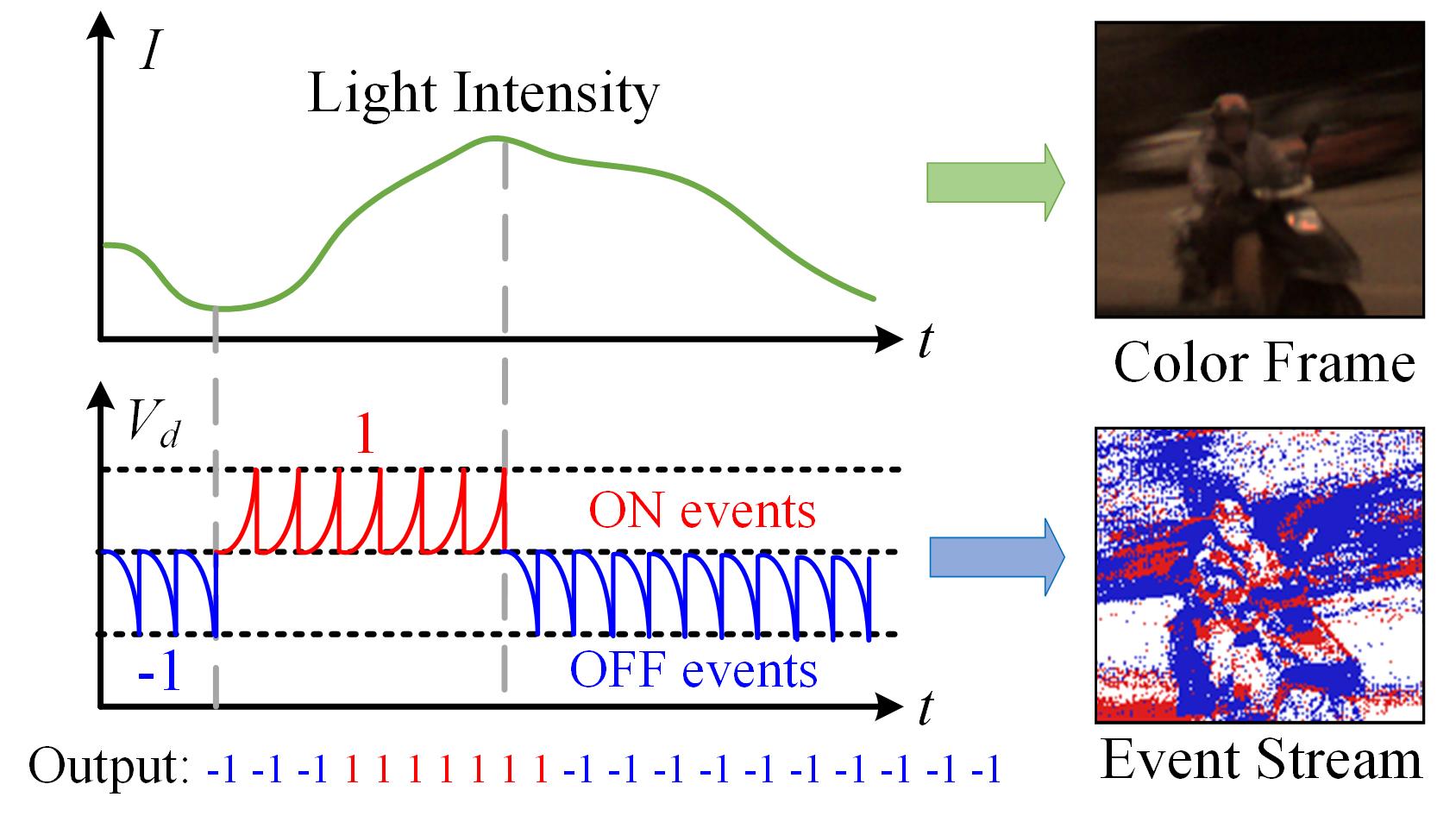}
\caption{Comparison between the imaging principles of the color frame and event stream.}     
\label{imagingPrinciple}
\end{figure}

Recently, the biologically inspired sensors (called event cameras), such as DAVIS~\cite{brandli2014240}, CeleX~\cite{chen2019celexV}, ATIS~\cite{posch2010qvga}, and PROPHESEE~\footnote{\url{https://www.prophesee.ai}}, drawing more and more attention of researchers. Different from RGB cameras which record light in a synchronous way (i.e., the video frame), the event cameras output events (or spikes) asynchronously which corresponds to the illumination variation. In another word, each pixel of event cameras independently records a binary value only when the light changes exceed a threshold. Events for the increase and decrease of illumination are called ON and OFF events respectively. Due to the unique sampling mechanism, the asynchronous events are spatially sparse but temporally dense. It is less affected by motion blur, therefore, is suitable for capturing fast-moving human actions, such as the magician's fast-moving palm, and movement recognition of sports players. It has a higher dynamic range (120 dB) and lower latency, which enables it to work well even in low illumination compared with standard RGB cameras. In addition, the storage and energy consumption are also significantly reduced~\cite{gallegoevent, wang2021visevent, Li2022vidardvsDet, zhu2022eventsnn, zhu2021neuspike}. Event streams highlight the contour information and protect personal privacy to a large extent. 
According to the aforementioned observation and thinking, we are inspired to address human activity recognition in the wild using event cameras. A comparison of the imaging principles of the color frame and event camera is illustrated in Fig.~\ref{imagingPrinciple}.

Although there are already several benchmark datasets proposed for classification~\cite{bi2020graph, amir2017low, li2017cifar10, serrano2015poker, kuehne2011hmdb, soomro2012ucf101, kliper2011action}. However, most of them are simulated/synthetic datasets that are transformed from RGB videos with the simulator. Some researchers attain the event data by recording the screen while displaying RGB videos.  Obviously, these datasets are hard to reflect the features of event cameras in real-world scenarios, especially fast-motion and low-light scenarios. ASL-DVS~\cite{bi2020graph} is proposed by Bi et al. which is consisted of $100,800$ samples but can only be used for hand gesture recognition with 24 classes. DvsGesture~\cite{amir2017low} is also limited by its scale and categories in the deep learning era. In addition, some datasets have become saturated in performance, for example, Wang et al.~\cite{wang2019steventclouds} already achieved $97.08\%$ on the DvsGesture~\cite{amir2017low} dataset. Therefore, the research community still has insistent demands for a large-scale HAR benchmark dataset recorded in the wild.

In this paper, we propose a large-scale benchmark dataset, termed HARDVS, to address the problem of the lack of real event data. Specifically, the HARDVS dataset contains more than 100K video clips recorded with a DAVIS346 camera, each of them lasting for about 5-10 seconds. It contains 300 categories of human activities in daily life, such as \emph{drinking, riding a bike, sitting down, washing hands}. The following factors are taken into account to make our data more diverse, including \emph{multi-views}, \emph{illuminations}, \emph{motion speed}, \emph{dynamic background}, \emph{occlusion}, \emph{flashing light}, \emph{photographic distance}. To the best of our knowledge, our proposed HARDVS is the first large-scale and challenging benchmark dataset for human activity recognition in the wild. A comparison between existing recognition datasets and our HARDVS is illustrated in Fig.~\ref{benchmarkCOMP}.

% Based on our newly proposed HARDVS dataset, we construct a novel hybrid framework for HAR, termed SCNNFormer. Specifically, it contains two major branches, i.e., the SCNN (spiking convolutional neural network) and Transformer module. The SCNN encodes the event streams along the simulation steps and achieves energy-efficient feature learning. The Transformer is adopted to capture the long-range relations in a parallel way. Also, we introduce the feature interactive learning layers to connect the dual branches and attain strong feature representation. The outputs will be concatenated and fed into the classification head for action prediction. An overview of our proposed SCNNFormer is illustrated in Fig.~\ref{framework}.  

Based on our newly proposed HARDVS dataset, we construct a novel event-based human action recognition framework, termed ESTF (Event-based Spatial-Temporal Transformer). As shown in Fig.~\ref{framework}, the ESTF transforms the event streams into spatial and temporal tokens and learns the dual features by employing SpatialFormer (SF) and TemporalFormer (TF) respectively.  Further, we propose a FusionFormer to realize the message passing between the spatial and temporal features. The aggregated representation is added with features of dual branches as the input for subsequent learning blocks, respectively. The outputs will be concatenated and fed into two MLP layers for the final action prediction.

To sum up, the contributions of this paper can be concluded as the following three aspects: 

$\bullet$ We propose a large-scale neuromorphic dataset for human activity recognition, termed HARDVS. It contains more than 100K samples with 300 classes, and fully reflects the challenging factors in the real world. To the best of our knowledge, it is the first large-scale realistic neuromorphic dataset for HAR. 

$\bullet$ We propose a novel Event-based Spatial-Temporal Transformer (ESTF) approach for human action recognition by exploiting spatial and temporal feature learning and fusing them  with Transformer networks. It is the first Transformer based spatial-temporal representation learning framework for event stream-based HAR. 

$\bullet$ We re-train and report the performance of multiple popular HAR algorithms, which provide extensive baselines for future works to compare on the HARDVS dataset. Extensive experiments on multiple event-based classification datasets fully demonstrate the effectiveness of our proposed ESTF approach.

\section{Related Work}
% In this section, we give a brief review of event camera-based HAR algorithms and benchmark datasets. More works about HAR and event cameras can be found in the following surveys~\cite{kong2018humanARSurvey, ahmad2021graph, gallegoevent}. 

\textbf{HAR with Event Sensors.}
Compared with RGB cameras, few researchers focus on event camera-based HAR~\cite{amir2017low, clady2017motion, chen2021novel, baby2017dynamic}. Arnon et al.~\cite{amir2017low} propose the first gesture recognition system based on TrueNorth neurosynaptic processor. Xavier et al.~\cite{clady2017motion} propose an event-based luminance-free feature for local corner detection and global gesture recognition. Chen et al.~\cite{chen2021novel} propose a hand gesture recognition system based on DVS and also design a wearable glove with a high-frequency active LED marker that fully exploits its properties. A retinomorphic event-driven representation (EDR) is proposed by Chen et al.~\cite{chen2019fast}, which can realize three important functions of the biological retina, i.e., the logarithmic transformation, ON/OFF pathways, and integration of multiple timescales. The authors of~\cite{lagorce2016hots} represent the recent temporal activity within a local spatial neighborhood, and utilize the rich temporal information provided by events to create contexts in the form of time-surfaces, termed HOTS, for the recognition task. Wu et al. first transform the event flow into images, then, predict and combine the human pose with event images for HAR~\cite{wu2020multipath}. Graph neural networks (GNN) and SNNs are also exploited for event-based recognition~\cite{george2020reservoir, mehr2019action, samadzadeh2020convsnn, li2018deepCNN, ceolini2020hand, panda2018learning, liu2020unsupervised, xing2020new, chen2020dyGCN, wang2021eventGNN}. Specifically, Chen et al.~\cite{chen2020dyGCN} treat the event flow as a 3D point cloud and use dynamic GNNs to learn the spatial-temporal features for gesture recognition. Wang et al.~\cite{wang2021eventGNN} adopt GNNs and CNNs for gait recognition. Xing et al. design a spiking convolutional recurrent neural network (SCRNN) architecture for event-based sequence~\cite{xing2020new}. According to our observations, these works are evaluated only on simple HAR datasets or simulated datasets. It is necessary and urgent to introduce a large-scale HAR dataset for current evaluation.

\textbf{Event Benchmark Datasets for HAR.} 
As shown in Table~\ref{datasetlist}, most of the existing event camera-based datasets for recognition are artificial datasets. Usually, the researchers display the RGB HAR datasets on a large screen and record the activity with neuromorphic sensors. For example, the N-Caltech101~\cite{orchard2015converting} and N-MNIST~\cite{orchard2015converting} are recorded with an ATIS camera which contains 101 and 10 classes, respectively. Bi et al.~\cite{bi2020graph} also transform popular HAR datasets into simulated event flow, including HMDB-DVS~\cite{bi2020graph, kuehne2011hmdb}, UCF-DVS~\cite{bi2020graph, soomro2012ucf101}, and ASLAN-DVS~\cite{kliper2011action}, which further expands the number of datasets available for HAR. However, these simulated event datasets hardly reflect the advantages of event cameras, such as low light, fast motion, etc. There are three realistic event datasets for classification, i.e., the DvsGesture~\cite{amir2017low}, N-CARS~\cite{sironi2018hats} and ASL-DVS~\cite{bi2020graph}, but these benchmarks are limited by their scale, categories, and scenes. Specifically, these datasets contain 11, 2, and 24 classes only, and also rarely take challenging factors like multi-view, motion, and glitter into consideration. Compared with existing datasets, our proposed HARDVS dataset is large-scale (100K samples) and category-wide (300 classes) for deep neural networks. Our sequences are recorded in the wild and fully reflect the features of the aforementioned attributes. We believe our proposed benchmark dataset greatly promotes the development of event-based HAR.

\begin{table*}[htp!]
\center
\scriptsize  
\caption{\textbf{Comparison of event datasets for human activity recognition.} M-VW, M-ILL, M-MO, DYB, OCC, and DR denotes multi-view, multi-illumination, multi-motion, dynamic background, occlusion, and duration of the action, respectively. Note that we only report these attributes of realistic DVS datasets for HAR.} \label{datasetlist}
\begin{tabular}{l|ccccccccccccc}
\hline \toprule [0.7 pt]
\textbf{Dataset}    &\textbf{Year}  &\textbf{Sensors}   &\textbf{Scale}   &\textbf{Class}  &\textbf{Resolution}  &\textbf{Real}  &\textbf{M-VW} &\textbf{M-ILL}  	&\textbf{M-MO}  &\textbf{DYB} &\textbf{OCC} &\textbf{DR} &\textbf{Link} \\ 
\hline 
\textbf{ASLAN-DVS}~\cite{bi2020graph, kliper2011action}   	 	&2011 	&DAVIS240c    &$3,697$ 		&432 			&$240 \times 180$ 		&\xmark	&-    &- 		&- 		&-		&- 	&-  	&\href{https://talhassner.github.io/home/projects/ASLAN/ASLAN-main.html}{URL} 				 \\
\textbf{MNIST-DVS}~\cite{serrano2015poker}   		 &2013   				&DAVIS128    &$30,000$ 		&10 		&$128\times128$ 			&\xmark  			&-    &- 		&- 		&-		&- 	&-  		&\href{http://www2.imse-cnm.csic.es/caviar/MNISTDVS.html}{URL} 				 \\
\textbf{N-Caltech101}~\cite{orchard2015converting}  &2015  &ATIS    &$8,709 $		&101 		&$302 \times 245$ 			&\xmark 	 &-    &- 		&- 		&-		&- 	&-  					&\href{https://www.garrickorchard.com/datasets}{URL} 				 \\
\textbf{N-MNIST}~\cite{orchard2015converting}   &2015  	&ATIS     &$70,000$ 		&10 		&$28\times28$ 			&\xmark  			&-    &- 		&- 		&-		&- 	&- 		&\href{https://www.garrickorchard.com/datasets}{URL} 				 \\
\textbf{CIFAR10-DVS}~\cite{li2017cifar10}   		 &2017 		&DAVIS128   &$10,000$ 		&10 		& $128\times128$			&\xmark  			&-    &- 		&- 		&-		&- 	&-  		&\href{https://figshare.com/s/d03a91081824536f12a8}{URL} 				 \\
\textbf{HMDB-DVS}~\cite{bi2020graph, kuehne2011hmdb}   	&2019  	&DAVIS240c    &$6,766$ 		&51 		&$240 \times 180$ 			&\xmark    &-    &- 		&- 		&-		&- 	&- 	&\href{https://serre-lab.clps.brown.edu/resource/hmdb-a-large-human-motion-database/}{URL} 				 \\
\textbf{UCF-DVS}~\cite{bi2020graph, soomro2012ucf101}   	&2019  	&DAVIS240c    &$13,320$  		&101 		&$240 \times 180$ 			&\xmark    &-    &- 		&- 		&-		&- 	&-  		&\href{https://www.crcv.ucf.edu/data/UCF101.php}{URL} 				 \\
\textbf{N-ImageNet}~\cite{kim2021nimagenet} &2021 	&Samsung-Gen3    &$1,781,167$   &1000   &$480 \times 640$   &\xmark  &-    &- 		&- 		&-		&- 	&- 	&\href{https://github.com/82magnolia/n_imagenet}{URL} \\ 
\textbf{ES-ImageNet}~\cite{lin2021esimagenet}  &2021 	&-    &$1,306,916$  &1000  		&$224 \times 224$  	 &\xmark  &-    &- 		&- 		&-		&- 	&- 	&\href{https://www.frontiersin.org/articles/10.3389/fnins.2021.726582/full?ref=https://githubhelp.com}{URL} 	  \\ 
\hline 
\textbf{DvsGesture}~\cite{amir2017low}    	&2017 		&DAVIS128   &$1,342$ 		&11 		&$128 \times 128$ 		&\cmark  			&\xmark    &\cmark 		&\xmark 		&\xmark			&\xmark  	& - 		&\href{https://research.ibm.com/dvsgesture/}{URL} 	  \\
\textbf{N-CARS}~\cite{sironi2018hats}   		 &2018  	&ATIS    &$24,029$ 		&2 		&$304 \times 240$  		&\cmark 	&\xmark    &\xmark 		&\xmark 		&\xmark			&\xmark 		 	&-  		&\href{https://www.prophesee.ai/2018/03/13/dataset-n-cars/}{URL} 				 \\
\textbf{ASL-DVS}~\cite{bi2020graph}   	 	&2019 	&DAVIS240    &$100,800$ 		&24 			&$240 \times 180$ 		&\cmark  	&\xmark    &\xmark  &\xmark 		&\xmark 	&\xmark  	&0.1s  	&\href{https://www.dropbox.com/sh/ibq0jsicatn7l6r/AACNrNELV56rs1YInMWUs9CAa?dl=0}{URL} 				 \\
\textbf{PAF}~\cite{miao2019neuromorphic}   	 	&2019 	&DAVIS346    &$450$ 		&10 			&$346 \times 260$ 		&\cmark  	&\xmark    &\xmark  &\xmark 		&\xmark 	&\xmark  	&5s  	&\href{https://github.com/CrystalMiaoshu/PAFBenchmark}{URL} 				 \\
\textbf{DailyAction}~\cite{LiuXTM021ijcai}   	 	&2021 	&DAVIS346    &$1,440$ 		&12 			&$346 \times 260$ 		&\cmark  	&\cmark    &\cmark  &\xmark 		&\xmark 	&\xmark  	&5s  	&\href{https://github.com/qianhuiliu/SNN-action-recognition}{URL} 				 \\
\hline 
\textbf{HARDVS (Ours)} &2022 		&DAVIS346    &$107,646$ 		&300		&$346 \times 260$ 		&\cmark  &\cmark    &\cmark 		&\cmark 		&\cmark			&\cmark 	  	&5s  	&\href{}{URL} 	  \\
\hline \toprule [0.7 pt]
\end{tabular}
\end{table*}

\begin{figure} 
\center
\includegraphics[width=3.3in]{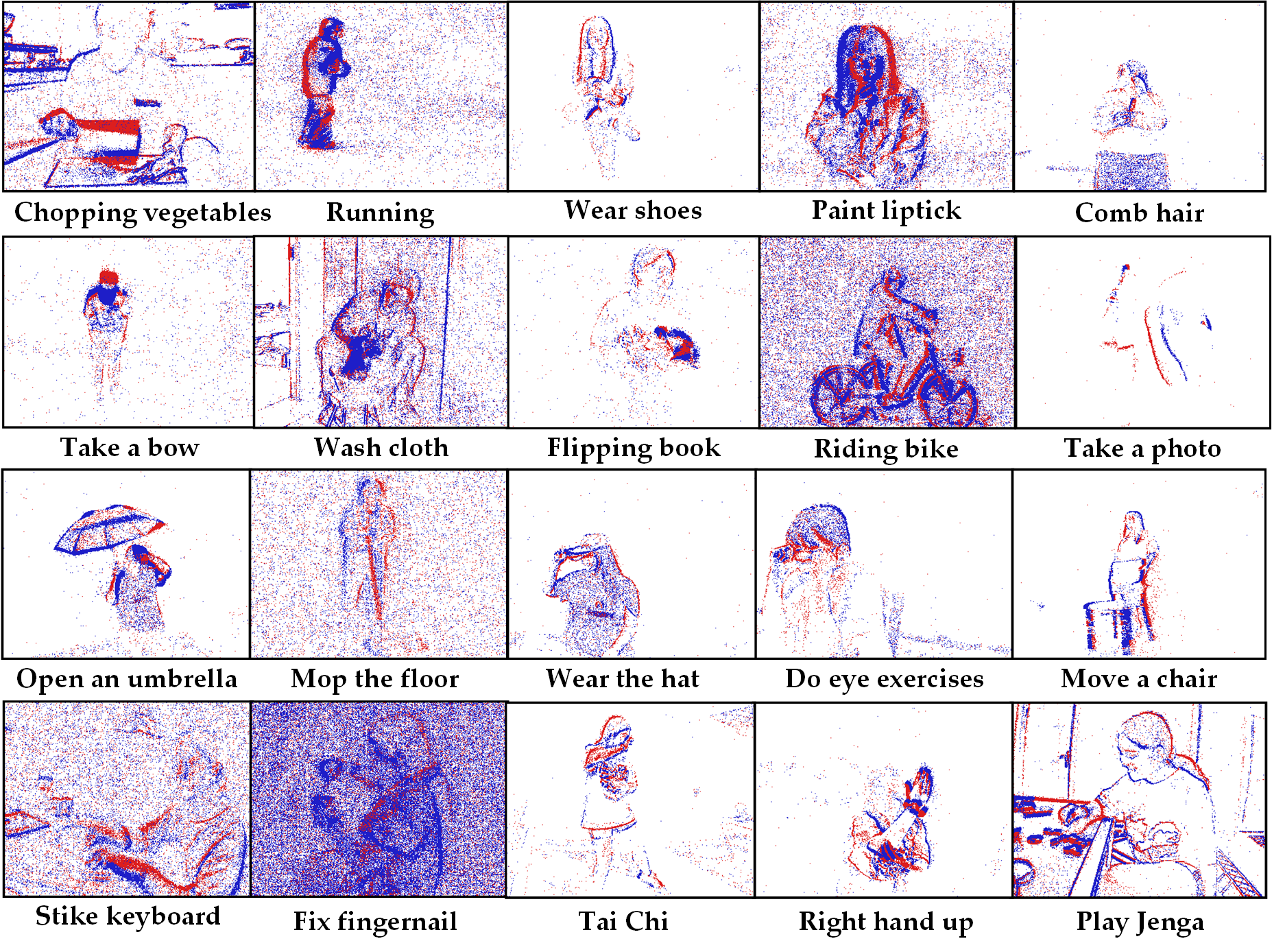}
\caption{Illustration of some representative samples of our proposed HARDVS dataset.}  
\label{HARDVS_samples}
\end{figure}

% \begin{figure*} 
% \center
% \includegraphics[width=7in]{figure/HARDVS_all_samples}
% \caption{Illustration of some representative samples of our proposed HARDVS dataset.}  
% \label{HARDVS_samples}
% \end{figure*} 

\section{HARDVS Benchmark Dataset} 
% In this section, we will first introduce the protocols of HARDVS dataset in Section \ref{Protocols}. Then, we will talk about the data collection and statistic analysis in Section \ref{dataCollection}. After that, we will introduce the baseline methods to evaluate on our HARDVS dataset in Section \ref{baselineConst}. 

\subsection{Protocols}  \label{Protocols}
We aim to provide a good platform for the training and evaluation of DVS-based human activity recognition. When constructing the HARDVS benchmark dataset, we obey the following protocols:

\textbf{1). Large-scale:} As we all know, large-scale datasets play a very important role in the deep learning era. In this work, we collect more than 100k DVS event sequences to meet the needs for large-scale training and evaluation of HAR. 
\textbf{2). Wide varieties:} Thousands of human activities can exist in the real world, but existing DVS-based HAR datasets only contain limited categories. Therefore, it is hard to fully reflect the classification and recognition ability of HAR algorithms. Our newly proposed HARDVS contains 300 classes which are several times larger than other DVS datasets.  
\textbf{3). Various challenges:} Our dataset considers multiple challenging factors which may influence the results of HAR with the DVS sensor. The detailed introductions can be found below: 
\emph{(a). Multi-view:} We collect different views of the same behavior to mimic practical applications, including front-, side-, horizontal-, top-down-, and bottom-up-views. 
\emph{(b). Multi-illumination:} High dynamic range is one of the most important features of DVS sensors, therefore, we collect the videos under scenarios with strong-, middle-, and low-light ($60\%$ of each category). Our dataset also contains many videos with \emph{glitter}, because we find that the DVS sensor is sensitive to flashing lights, especially at the night. 
\emph{(c). Multi-motion:} We also highlight the features of DVS sensors by recording many actions with various motion speeds, such as slow-, moderate-, and high-speed. 
\emph{(d). Dynamic background:} As it is relatively easy to recognize actions without background objects, i.e., stationary DVS camera, we also collect many actions with a dynamic moving camera to make our dataset challenging enough. 
\emph{(e). Occlusion:} In the real world, human action can be occluded commonly. Thus, we also add occlusion issues into the HARDVS dataset with hand or other things. 
\textbf{4). Different capture distance:} The HARDVS dataset is collected under various distances, i.e., 1-2, 3-4, and more than 5 meters. 
\textbf{5). Long-term:} Most of the existing DVS-based HAR datasets are microsecond-level, in contrast, each video in our HARDVS dataset lasts for about 5 seconds.  
\textbf{6). Dual-modality:} The DAVIS346 camera can output both RGB frames and event flow, therefore, our dataset can also be used for HAR by fusing video frames and events. In this work, we focus on HAR with DVS only, but the RGB frames will also be released to support the research on dual-modality fusing based HAR.

\subsection{Data Collection and Statistic Analysis} \label{dataCollection}
The HARDVS dataset is collected with a DAVIS346 camera whose resolution is $346 \times 260$. We take the aforementioned protocols in mind when recording videos. Therefore, our dataset fully reflects the unique features of DVS sensors in challenging scenarios, such as low-illumination, high-speed, clutter background, etc. The main characters are also diverse, generally speaking, there is a total of five persons involved in the data collection stage.  

From a statistical perspective, our dataset contains a total of $107,646$ video sequences and 300 classes of common human activities. We split $60\%, 10\%$, and $30\%$ of each category for training, validating, and testing, respectively. Totally, the number of videos in the training, validating, and testing subset is $64526 | 10734 | 32386$, respectively. A direct comparison with existing classification benchmark datasets can be found in Table~\ref{datasetlist} and Fig.~\ref{benchmarkCOMP}. 
% We also give a visualization of video numbers of each class in our HARDVS dataset, as shown in Fig.~\ref{distributionHARDVS}. 
With the aforementioned characteristics, we believe our HARDVS dataset will be a better evaluation platform for the neuromorphic classification problem, especially for the human activity recognition task.

\section{Methodology}

\subsection{Overview} 
\begin{figure*}
\center
\includegraphics[width=7in]{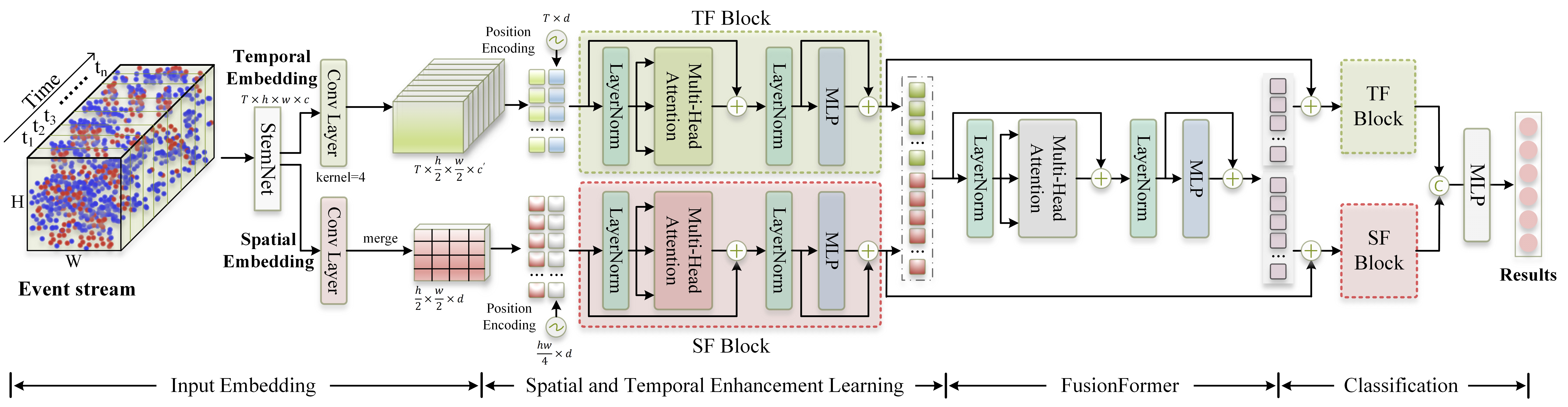}
\caption{\textbf{An overview of our proposed ESTF framework for event-based human action recognition.} It transforms the event streams into spatial and temporal tokens and learns the dual features using multi-head self-attention layers. Further, a FusionFormer is proposed to realize message passing between the spatial and temporal features. The aggregated features are added with dual features as the input for subsequent TF and SF blocks, respectively. The outputs will be concatenated and fed into MLP layers for action prediction. 
}
\label{framework}
\end{figure*}

In this section, we devise a new Event-based Spatial-Temporal Transformer (ESTF) approach for event-stream data learning. 
As shown in Fig.~\ref{framework}, the proposed ESTF architecture contains three main learning modules, i.e., i) Initial Spatial and Temporal Embedding, ii) Spatial and Temporal Enhancement Learning, and iii) Spatial-Temporal Fusion Transformer. Specifically, 
given the input event-stream data, 
we first extract the initial spatial and temporal embeddings respectively. 
Then, a Spatial and Temporal Feature Enhancement Learning module is devised to further enrich the  event-stream data representations by deeply capturing both \textbf{spatial correlation} and \textbf{temporal dependence}  of 
event stream. Finally, an effective Fusion Transformer (FusionFormer) block is designed to integrate the spatial and temporal cues together for the final feature representation. 
The details of these modules are introduced below. 

\subsection{Initial Spatial and Temporal Embedding} 
Different from visible sensors which capture a global image at each time, the event cameras asynchronously capture the intensity variations in the log-scale. That is, each pixel outputs a discrete event (or spike) independently when the visual changing exceeds a pre-defined threshold. 
Usually, we use a 4-tuple $\{x, y, t, p\}$ to represent the discrete event of a pixel captured with DVS, where $x, y$ are spatial coordinates, $t$ is timestamp, and $p\in \{1,-1\}$ is the polarity of brightness variation. 
Following previous works~\cite{wang2019evGait, zhu2018evflownet, fang2021snnresnet, yao2021TASNN}, 
we first transform the asynchronous event flow into the synchronous \emph{event images} by stacking the events in a time interval based on the exposure time. Let $\mathcal{E}=\{E_1, E_2\cdots E_T\}\in \mathbb{R}^{H\times W \times T}$ be the collection of the sampled input event frames. In our experiments, we set $T =8$, as used in works~\cite{tran2015c3d}. 
For each event frame $E_t$, we adopt StemNet~\cite{he2016resnet} to extract an initial CNN feature descriptor for it and denote 
$\mathcal{X}=\{\mathcal{X}_1, \mathcal{X}_2\cdots \mathcal{X}_T\}\in \mathbb{R}^{T\times h\times w\times c}$ as the collection of $T$ event frames. 
Based on it, we respectively  extract spatial and temporal embeddings. 
To be specific, for temporal branch, we adopt a convolution layer to reduce the feature size to obtain 
${\mathcal{X}^t}\in \mathbb{R}^{T\times \frac{h}{2}\times \frac{w}{2}\times c'}$ and reshape it to the matrix form as
$X^t\in \mathbb{R}^{T\times d}$ where $d=\frac{h}{2}\times \frac{w}{2}\times c'$. 
For spatial branch, 
we first adopt a convolution layer 
to resize the features 
$\mathcal{X}$ to 
${\mathcal{X}^s}\in \mathbb{R}^{T\times \frac{h}{2}\times \frac{w}{2}\times d}$. 
Then, we  conduct the merging/summation operation  on the time dimension and reshape it to the matrix form 
 ${X}^s \in \mathbb{R}^{N\times d}$ where $N=\frac{hw}{4}$. 
Hence, both spatial and temporal embeddings have the same $d$-dim feature descriptors. % feature dimension. 
% as 
%the feature collection of patches where  $M=m\times n$ is the number of all patches. 

%To fully utilize the benefits of CNN, previous works usually transform the asynchronous event flow into synchronous \emph{event image} by stacking the events in a fixed time interval. In this work, we also adopt such transformation to get the event images. Without loss of generality, we color the positive and negative events with red and blue, therefore, we can directly train the existing HAR algorithms developed for RGB video frames, as shown in Fig. \ref{HARDVSsamples}. Based on such event image representation, we build a novel SCNNFormer framework for HAR.   

\subsection{Spatial and Temporal Enhancement Learning}
Based on the above initial spatial embeddings $X^s\in \mathbb{R}^{N\times d}$ and temporal embeddings $X^t\in \mathbb{R}^{T\times d}$, we then 
devise our  Spatial and Temporal Enhancement Learning (STEL) module to further enrich their representations. 
The proposed  STEL module involves two blocks, i.e., Spatial Transformer (SF) block, and Temporal Transformer (TF) block, which respectively capture the spatial correlations and {temporal dependences}  of 
event data to learn context enriched representations. 
The SF block includes multi-head self-attention (MSA) and MLP module with a Layernorm (LN)  used between two modules.  
A  residual connection is also employed, as shown in Fig.~\ref{framework}. To be specific, given spatial embeddings $X^s\in \mathbb{R}^{N\times d}$, we first incorporate  the position encoding~\cite{dosovitskiy2020ViT} to obtain  $\bar{X}^s\in \mathbb{R}^{N\times d}$ which represents 
% Let $X^s\in \mathbb{R}^{N\times d}$ denotes 
$N$ number of the input tokens with $d$-dim feature descriptor. % is the feature dimension of each token. 
Then, the outputs of SF block are summarized as follows,  
\begin{align}
\label{sffeature}
& Y^s = LN(\bar{X}^s + MSA(LN(\bar{X}^s)))\\
& \widetilde{X}^s = Y^s+MLP(Y^s)
\end{align}
In contrast to input $\bar{X}^s$, the output $\widetilde{X}^s$ provides the spatial-aware enhanced representations by employing the MSA mechanism to model the spatial relationships of different event patches. 
Similarly, given $\bar{X}^t\in \mathbb{R}^{T\times d}$ representing $T$  temporal tokens with position encoding, the outputs of TF block are summarized as follows, 
\begin{align}
\label{tffeature}
& Y^t = LN(\bar{X}^t + MSA(LN(\bar{X}^t)))\\
& \widetilde{X}^t = Y^t+MLP(Y^t)
\end{align}
Compared with the input $\bar{X}^t$, the outputs $\widetilde{X}^t\in \mathbb{R}^{T\times d}$ provide a temporal-context enhanced representations for $T$ number of frame tokens thanks to the MSA mechanism to model the  dependencies of different event frames.

\subsection{Fusion Transformer} 

In order to conduct the interaction between the above 
ST and TF blocks and learn a unified spatio-temporal contextual data representations, we also design a Fusion Transformer (FusionF) module. 
To be specific, 
let $\widetilde{X}^s$ and $\widetilde{X}^t$ denote the 
outputs of previous SF and TF blocks respectively. 
We first 
collect the $N$ spatial and $T$ temporal tokens together and 
 feed them to a unified Transformer block which includes 
multi-head self-attention (MSA) and MLP submodule, i.e.,  
\begin{align}
& Z = [\widetilde{X}^t, \widetilde{X}^s] \in \mathbb{R}^{(T+N)\times c} \\
& {Y} = LN(Z + MSA(LN(Z)))\\
&\widetilde{Z} = Z+Y+MLP(Y)
\end{align}
%
% $\phi(F, \Theta)$
Afterword, 
we split $\widetilde{Z}$ into 
$\{\widetilde{Z}^s,\widetilde{Z}^t\}$ where $\widetilde{Z}^s\in \mathbb{R}^{N\times d} $ and $\widetilde{Z}^t\in \mathbb{R}^{T\times d}$ and further employ the above SF (Eqs.(1,2)) and TF (Eqs.(3,4)) block  to  respectively enhance their representations as follows, 
\begin{align}
F^s =SF (\widetilde{Z}^s), F^t =TF (\widetilde{Z}^t)
\end{align}
Finally, we concatenate both $F^s$ and $F^t$ together and reshape the concatenated features to the vector form. After that, we utilize a two-layer MLP to output the final class label prediction, as shown in Fig.~\ref{framework}.

\subsection{Loss Function}  
Our proposed ESTF framework can be optimized in an end-to-end way. The standard cross-entropy loss function is adopted to measure the distance between our model prediction and ground truth: 
\begin{align}
Loss =-\frac{1}{B}\sum_{b=1}^{B}\sum_{n=1}^{N}Y_{bn}logP_{bn} 
\end{align}
where $B$ denotes the batch size, $N$ denotes the number of event classes. $Y$ and $P$ represent the ground truth and predicted class labels of the event sample, respectively.

% which is applied to minimize the error between the prediction and ground-truth event label.

\section{Experiments} 

\subsection{Dataset and Evaluation Metrics} 
In this work, three datasets are adopted for the evaluation of our proposed model, including \textbf{N-Caltech101}~\cite{orchard2015converting}, \textbf{ASL-DVS}~\cite{bi2020graph}, and our newly proposed \textbf{HARDVS}. More details about these datasets can be found in Table~\ref{datasetlist}. The widely used \textbf{top-1} and \textbf{top-5 accuracy} are adopted as evaluation metrics.

\subsection{Implementation Details} 
Given the event streams, we stack them into image-like representations to make full use of CNN. More in detail, the time window is set based on the exposure time of color frames, when generating the event images. The batch size is 60, and the initial learning rate is 0.01, which is reduced to 10\% of the original every 15 epochs. The stochastic gradient descent (SGD)~\cite{sutskever2013SGD} is selected as the optimizer to train our network. Our code is implemented based on Python 3.8, PyTorch 1.10.2+cu113~\cite{paszke2019pytorch}, on a server with RTX3090. The source code and pre-trained models will be released to help other researchers reproduce our experimental results.

\subsection{Comparison with SOTA Algorithms} 

\textbf{Results on N-Caltech101}~\cite{orchard2015converting}. 
As shown in Table~\ref{Caltech101Results}, our proposed method achieves 0.832 on the top-1 accuracy metric which is significantly better than the compared models by a large margin. For example, the VMV-GCN achieves 0.778 on this benchmark dataset which ranks second place, meanwhile, our model outperforms it by up to $5.4\%$. The M-LSTM is an adaptive event representation learning model which obtained 0.738 only on this dataset. EV-VGCNN is a graph neural network based model which obtains 0.748 and is also worse than ours. These experimental results fully demonstrate the effectiveness of our proposed spatial-temporal feature learning for event-based pattern recognition.

\begin{table}
\center
\scriptsize  
\caption{Results on N-Caltech101~\cite{orchard2015converting} Dataset.} 
\label{Caltech101Results} 
\setlength\tabcolsep{3.3pt}
\begin{tabular}{ccccccc} 		%% \xmark   \cmark   
\hline \toprule [0.5 pt] 
\textbf{EventNet}   &\textbf{Gabor-SNN}     &\textbf{RG-CNNs}     &\textbf{VMV-GCN}     &\textbf{EV-VGCNN}     &\textbf{EST}     	  		 \\
0.425    &0.196     &0.657     &\textcolor{blue}{\textbf{0.778}}     &0.748     &0.753     	  		 \\
\hline
\textbf{ResNet-50}  &\textbf{MVF-Net}     &\textbf{M-LSTM}     &\textbf{AMAE}     &\textbf{HATS}     &\textbf{Ours}     	  		 \\ 
0.637    &0.687     &0.738     &0.694     &0.642 &\textcolor{red}{\textbf{0.832}}     	  		 \\
\hline \toprule [0.5 pt] 
\end{tabular}
\end{table}

\textbf{Results on ASL-DVS}~\cite{bi2020graph}.
As shown in Table~\ref{ASLDVSResults}, the performance on this dataset is already close to saturation and most of the compared models achieve more than 0.95+ on the top-1 accuracy, including EST~\cite{gehrig2019EST} (0.979), AMAE~\cite{deng2020amae} (0.984), M-LSTM~\cite{cannici2020mlstm} (0.980), MVF-Net~\cite{deng2021mvfnet} (0.971). Note that, the VMV-GCN~\cite{xie2022vmvgcn} achieves 0.989 on this benchmark dataset which ranks the second place. It is very hard to beat these models. Thanks to our proposed spatial-temporal feature learning and fusion modules, we set new state-of-the-art performance on this dataset, i.e., 0.999 on the top-1 accuracy. Therefore, we can draw the conclusion that our method almost completely solves the simple gesture recognition problem defined in the ASL-DVS~\cite{bi2020graph}.

\begin{table}
\center
\scriptsize   
\caption{Results on the ASL-DVS~\cite{bi2020graph} dataset.} 
\label{ASLDVSResults} 
\begin{tabular}{ccccccccccccccc} 		%% \xmark   \cmark   
\hline \toprule [0.5 pt] 
\textbf{EST}   &\textbf{AMAE}     &\textbf{M-LSTM}     &\textbf{MVF-Net}     &\textbf{ResNet-50}  \\  
0.979   & 0.984     &0.980     &0.971     &0.886     \\ 
\hline 
\textbf{EventNet}     &\textbf{RG-CNNs}     &\textbf{\makecell[c]{EV-VGCNN}}     &\textbf{\makecell[c]{VMV-GCN}}     &\textbf{Ours}     	  		  		 \\
0.833  &0.901     &0.983     &\textcolor{blue}{\textbf{0.989}}     &\textcolor{red}{\textbf{0.999}}	 \\
\hline \toprule [0.5 pt] 
\end{tabular}
\end{table}

\textbf{Results on HARDVS.} 
From the experimental results reported in the ASL-DVS~\cite{bi2020graph} and N-Caltech101~\cite{orchard2015converting}, we can find that existing event based recognition datasets are almost saturated. The newly proposed HARDVS dataset can bridge this gap and further boost the development of event based human action recognition. As shown in Table~\ref{hardvsResults}, we re-training and testing multiple state-of-the-art models for future works to compare on the HARDVS benchmark dataset, including C3D~\cite{tran2015c3d}, R2Plus1D~\cite{tran2018R2Plus1D}, TSM~\cite{song2019TSM}, ACTION-Net~\cite{wang2021actionnet}, TAM~\cite{liu2021tam}, Video-SwinTrans~\cite{liu2021videoSwin}, TimeSformer~\cite{bertasius2021TimeSformer}, SlowFast~\cite{feichtenhofer2019slowfast}. It is easy to find that these popular and strong recognition models still perform poorly on our newly proposed HARDVS dataset. To be specific, the R2Plus1D~\cite{tran2018R2Plus1D}, ACTION-Net~\cite{wang2021actionnet}, and SlowFast~\cite{feichtenhofer2019slowfast} only achieves $49.06 | 56.43$, $46.85 | 56.19$, and $46.54 | 54.76$ on the top-1 and top-5 accuracy respectively. The recently proposed TAM~\cite{liu2021tam} (ICCV-2021), Video-SwinTrans~\cite{liu2021videoSwin} (CVPR-2022), TimeSformer~\cite{bertasius2021TimeSformer} (ICML 2021) also obtains $50.41 | 57.99$, $51.91 | 59.11$, and $50.77 | 58.70$ on the two metrics respectively. Compared with these models, our proposed spatial-temporal feature learning and fusion modules perform comparable or even better than these SOTA models, i.e., $51.22 | 57.53$. All in all, our proposed model is effective for event based human action recognition task and may be a good baseline for future works to compare.

\begin{table} 
\scriptsize     
\center 
\caption{Results on the newly proposed HARDVS dataset.}  
\label{hardvsResults}
\setlength\tabcolsep{3.3pt}
\begin{tabular}{c|l|l|c|ccccccc} 
\hline \toprule [0.7 pt]
\textbf{No.} &\textbf{Algorithm} &\textbf{Publish}  &\textbf{Backbone} &\textbf{Top1}  &\textbf{Top5} \\
\hline 
01 & ResNet18~\cite{he2016resnet} &CVPR-2016    &ResNet18      &49.20  &56.09     \\
\hline 
02 & C3D~\cite{tran2015c3d} &ICCV-2015   &CNN   &50.52  &56.14     \\
\hline 	
03 & R2Plus1D \cite{tran2018R2Plus1D} &CVPR-2018 &ResNet-34      &49.06  &56.43     \\
\hline 	
04 & TSM ~\cite{lin2019tsm} &ICCV-2019     &ResNet-50       &52.63  &60.56    \\
\hline 	
05 & ACTION-Net~\cite{wang2021actionnet} &CVPR-2021     &ResNet-50       &46.85  &56.19    \\
\hline 	
06 & TAM \cite{liu2021tam} 	&ICCV-2021     &ResNet-50       &50.41  &57.99    \\
\hline 
07 &  Video-SwinTrans \cite{liu2021videoSwin}   &CVPR-2022     &Swin Transformer      &51.91  &59.11    \\
\hline 
08 & TimeSformer \cite{bertasius2021TimeSformer}  &ICML-2021     &VIT      &50.77  &58.70    \\
\hline 
09 & SlowFast  \cite{feichtenhofer2019slowfast}  &ICCV-2019     &ResNet-50       &46.54  &54.76    \\
\hline 
10 & X3D  \cite{feichtenhofer2020x3d}  &CVPR-2020     &ResNet       &45.82  &52.33    \\
\hline 
11 & ESTF (Ours) &-       &ResNet18       &51.22  &57.53    \\
\hline \toprule [0.7 pt]
\end{tabular}   
\end{table}

\subsection{Ablation Study} 

To help researchers better understand our proposed module, in this subsection, we conduct extensive experiments to analyze the contributions of each key component and the influence of different settings for our model.

\textbf{Component Analysis.} 
As shown in Table~\ref{CAResults}, three main modules are analyzed on the N-Caltech101 dataset, including SpatialFormer (SF), TemporalFormer (TF), and FusionFormer. We can find that our baseline method ResNet18~\cite{he2016resnet} achieves 72.14 on the top-1 accuracy metric. When introducing the TemporalFormer (TF) into the recognition framework, the overall performance can be significantly improved by $+9.4$, and achieves 81.54. When the SpatialFormer (SF) is adopted for long-range global feature relation mining, the recognition results can be enhanced to 80.47, and the improvement is up to $+8.33$. When both modules are all utilized for joint spatial-temporal feature learning, a better result can be obtained, i.e., 82.89. If the FusionFormer is adopted to achieve interactive feature learning and information propagation between the spatial and temporal Transformer branches, the best results can be achieved, i.e., 83.17 on the top-1 accuracy. Based on the experimental analysis for Table~\ref{CAResults} and Table~\ref{Caltech101Results}, we can draw the conclusion that our proposed modules all contribute to final recognition results.

\begin{table}[!htp]
\center
\small   
\caption{Component Analysis on the N-Caltech101 Dataset.} \label{CAResults} 
\begin{tabular}{c|cccc|cc} 		%% \xmark   \cmark   
\hline \toprule [0.5 pt] 
\textbf{No.}  & \textbf{ResNet} &\textbf{TF} &\textbf{SF}  &\textbf{FusionFormer}  &\textbf{Accuracy}  \\
\hline 
1 &\cmark   &          &         &         &72.14   \\
\rowcolor{mygray}
2 &\cmark   &\cmark    &         &         &81.54   \\
3 &\cmark   &    &\cmark         &         &80.47   \\
\rowcolor{mygray}
4 &\cmark   &\cmark    &\cmark   &         &82.89    \\
5 &\cmark   &\cmark    &\cmark   &\cmark   &83.17    \\  
\hline \toprule [0.5 pt]
\end{tabular}
\end{table}

\begin{figure} 
\center
\includegraphics[width=3.3in]{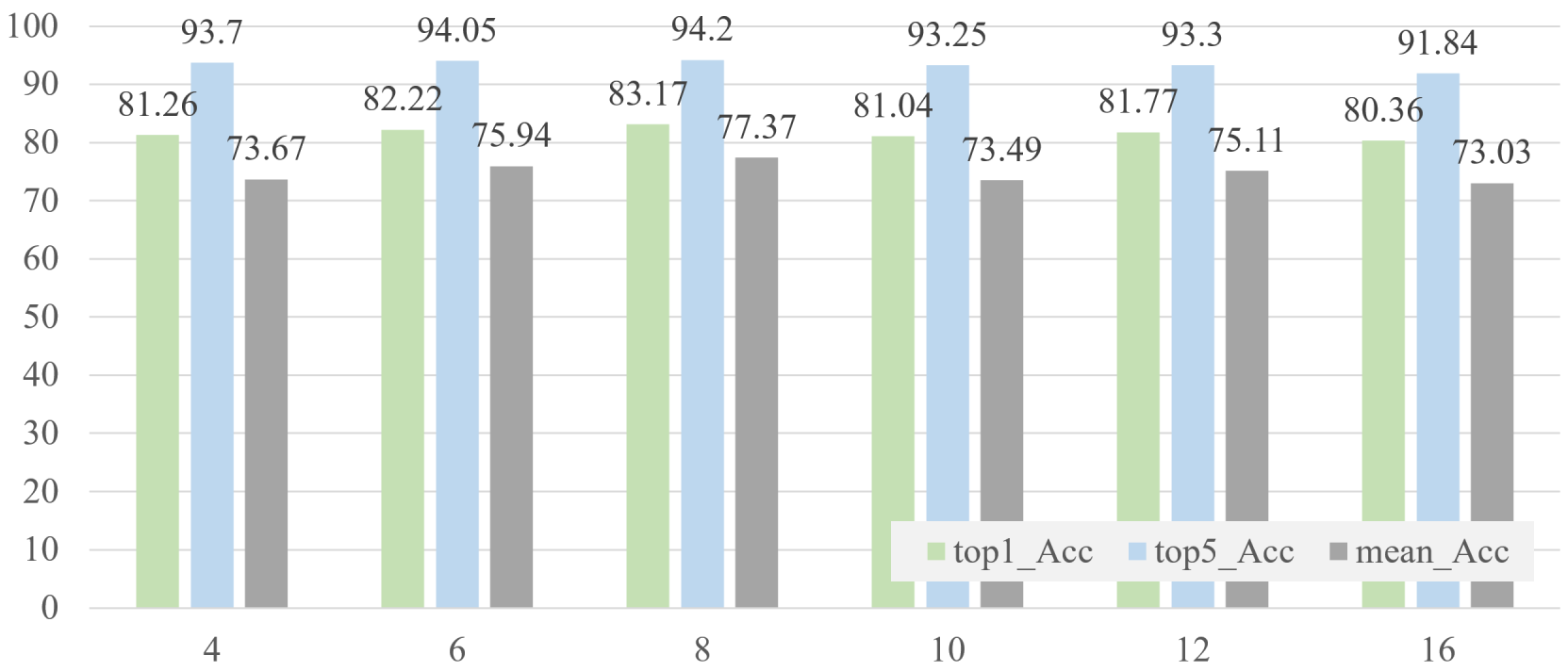}
\caption{Experimental results of different input frames.} 
\label{Param_InputFrames}
\end{figure} 

\textbf{Analysis on Number of Input Frames.} 
In this paper, we transform the event streams into an image-like representation for classification. In our experiments, 8 frames are adopted for the evaluation of our model. Actually, various event frames can be obtained with different intervals of the time windows. In this part, we test our model with 4, 6, 8, 10, 12, and 16 frames on the N-Caltech101 dataset and report the results in Fig.~\ref{Param_InputFrames}. It is easy to find that the mean accuracy is 73.67, 75.94, 77.37, 73.49, 75.11, and 73.03, correspondingly, and the highest mean accuracy can be obtained when 8 frames are adopted. For the decrease in accuracy when the frames are larger than 8, we think this may be caused by the fact that the event streams are partitioned into more frames and each frame will be more sparse. Therefore, this will lead to sparse edge information which is very important for recognition.

\begin{figure} 
\center
\includegraphics[width=3.3in]{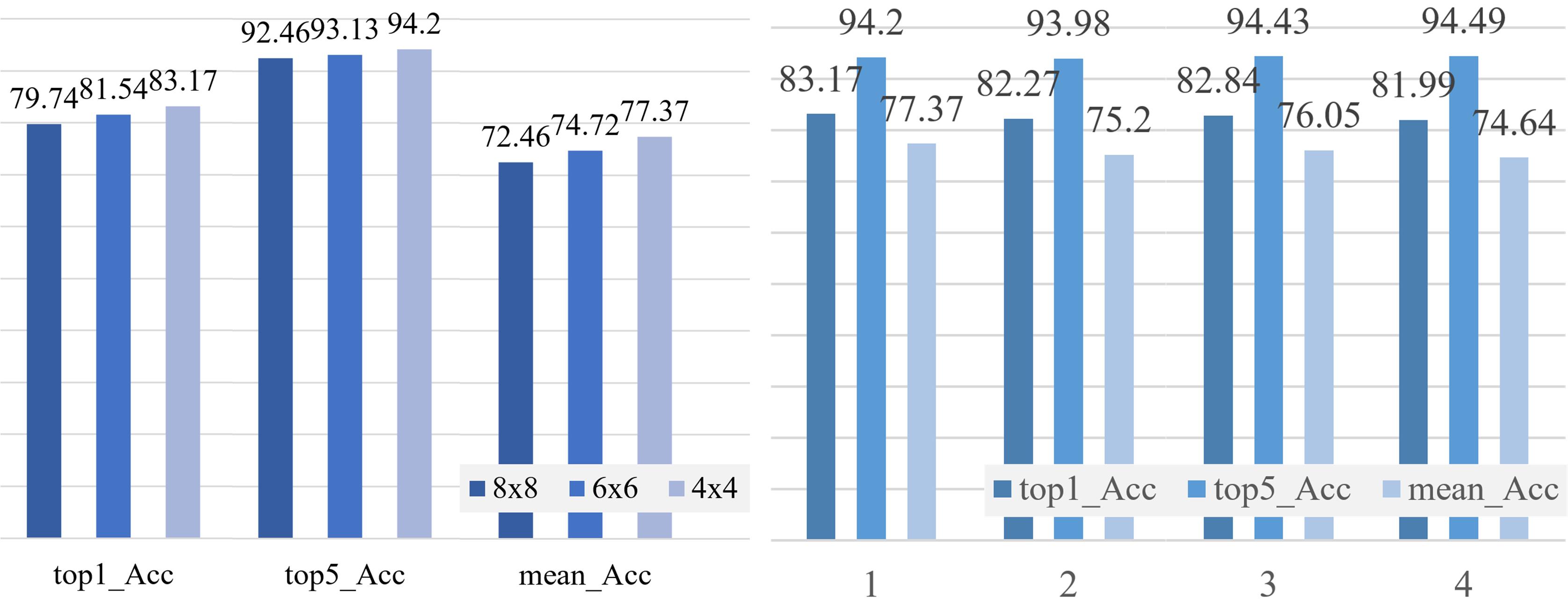}
\caption{Experimental results of different partition patches (left) and Transformer layers (right).}    
\label{Param_NumberLayers}
\end{figure}

\begin{figure*}[!htp]
\center
\includegraphics[width=7in]{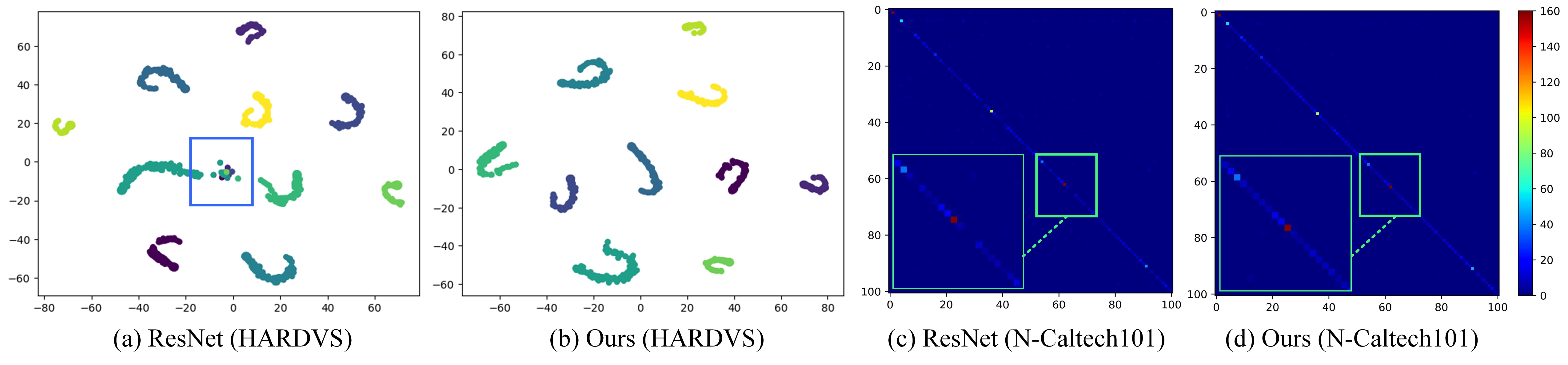}
\caption{Visualization of feature distribution of our baseline and newly proposed ESTF on HARDVS dataset (a, b) and confusion matrix of baseline ResNet and our model on N-Caltech101 dataset (c, d). Best viewed by zooming in.}
\label{featureconfusionMarixVIS}
\end{figure*}

\begin{figure*}[!htp]
\center
\includegraphics[width=7in]{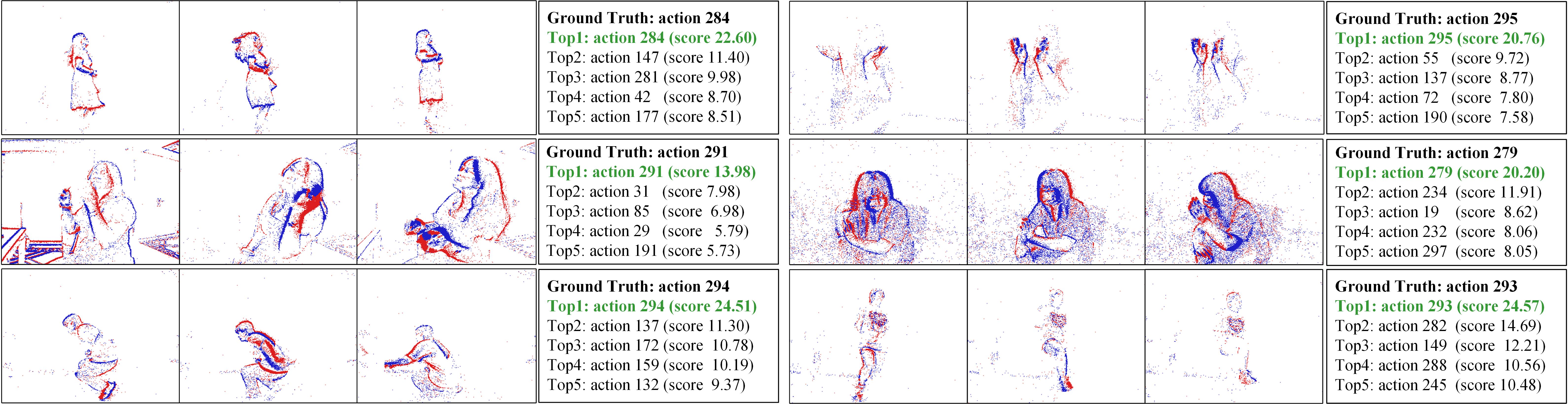}
\caption{Visualization of the top-5 predicted actions using our model.}
\label{framesVIS}
\end{figure*}

\textbf{Analysis on Split Patches of Spatial Data.} 
In this paper, the spatial features are partitioned into non-overlapped patches. We test multiple scales in this subsection, including $8 \times 8$, $6 \times 6$, and $4 \times 4$. As illustrated in Fig.~\ref{Param_NumberLayers} (left), the best performance can be obtained when $4 \times 4$ is adopted, i.e., 83.17, 94.20, and 77.37 on the top-1, top-5, and mean accuracy respectively.

\textbf{Analysis on Layers of Transformer Layers.} 
As we all know, the self-attention or Transformer layers can be stacked multiple times for more accurate recognition, as validated in many works. In this experiment, we also test different Transformer layers to check their influence on our model. As shown in Fig.~\ref{Param_NumberLayers} (right), four different settings are tested, i.e., 1, 2, 3, and 4 layers, and the corresponding mean accuracy is 77.37, 75.20, 76.05, and 74.64. We can find that higher recognition results can be obtained when the Transformer is set as 1 to 3 layers. Maybe a larger dataset is needed to train deeper Transformer layers.

\textbf{Model Parameters and Running Efficiency.} 
The storage space occupied by our checkpoint is 377.34 MB and the number of parameters is 46.71 M. The MAC score is 17.62 G tested using toolkit \emph{ptflops} \footnote{\url{https://pypi.org/project/ptflops/}}. Our model spends 25 ms for each video (8 frames used) in our proposed HARDVS dataset.

\begin{figure}[!htp]
\center
\includegraphics[width=3.3in]{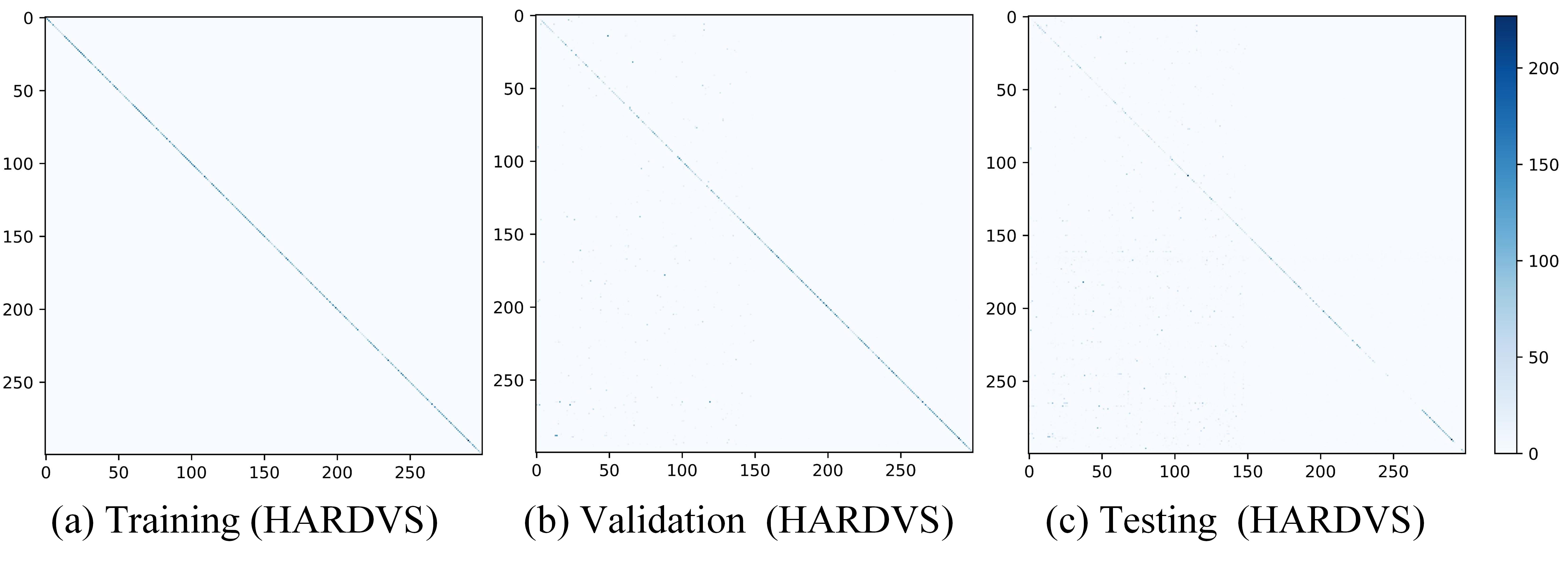}
\caption{Visualization of confusion matrix on the HARDVS dataset.}
\label{confusionMatrixVIS2}
\end{figure}

\subsection{Visualization}
In the previous subsections, we conduct extensive experiments to validate the effectiveness of our model from a quantitative point of view. In this part, we resort to the visualization to help the readers better understand our proposed model.

\textbf{Feature Visualization \& Confusion Matrix.~}  
As shown in Fig.~\ref{featureconfusionMarixVIS} (a, b), we select 10 classes of actions defined in the HARDVS dataset and visualize the features by projecting them into 2D plane using \emph{tSNE} toolkit \footnote{\url{https://github.com/mxl1990/tsne-pytorch
}}. It is easy to find that partial data samples are not discriminated well using the baseline ResNet18, such as the regions highlighted in blue bounding box. In contrast, our proposed ESTF model achieves a better feature representation learning and more of the categories are classified well. 
For the confusion matrix on N-Caltech101 dataset, as shown in Fig.~\ref{featureconfusionMarixVIS} (c, d), we can find that our proposed ESTF achieves significant improvement compared with our baseline ResNet18. 
All in all, we can draw the conclusion that our proposed spatial-temporal feature learning module works well for event based action recognition.

\textbf{Confusion Matrix.~} As shown in Fig.~\ref{confusionMatrixVIS2}, we visualize the confusion matrix of our model based on the results predicted in the training, validation, and testing phase, respectively. One can note that our model achieves better results in the training phase, but the overall performance in the testing phase is still weak. This demonstrate that our proposed HARDVS dataset is challenging and there is still plenty of room for further improvement.

\textbf{Recognition Results.~}
As shown in Fig.~\ref{framesVIS}, we provide the top-5 predicted actions and corresponding confidence scores. The ground truth and top-1 results are highlighted in \textbf{black} and \textbf{\textcolor{SeaGreen4}{green}}. It is easy to find that our model can predict the human activities accurately.

% \textbf{Limitation Analysis.}

\section{Conclusion} 
In this paper, we propose a large-scale benchmark dataset for event-based human action recognition, termed HARDVS. It contains 300 categories of human activities and more than 100K event sequences captured from DAVIS346 camera. These videos reflect various views, illuminations, motions, dynamic backgrounds, occlusion, etc. More than 10 popular and recent classification models are evaluated for future works to compare. In addition, we also propose a novel Event-based Spatial-Temporal Transformer (short for ESTF) that conducts spatial-temporal enhanced learning and fusion for accurate action recognition. Extensive experiments on multiple benchmark datasets validated the effectiveness of our proposed framework. It sets the new SOTA performances on N-Caltech101 and ALS-DVS datasets. We hope the proposed dataset and baseline approach will boost the further development of event camera based human action recognition. 
In our future works, we will consider combining the color frames and event streams together for high-performance action recognition.

%%%%%%%%% REFERENCES
\small{
\bibliographystyle{ieee_fullname}
\bibliography{reference}

\begin{thebibliography}{10}\itemsep=-1pt

\bibitem{ahmad2021graph}
Tasweer Ahmad, Lianwen Jin, Xin Zhang, Luojun Lin, and Guozhi Tang.
\newblock Graph convolutional neural network for action recognition: A
  comprehensive survey.
\newblock {\em IEEE Transactions on Artificial Intelligence}, 2021.

\bibitem{amir2017low}
Arnon Amir, Brian Taba, David Berg, Timothy Melano, Jeffrey McKinstry, Carmelo
  Di~Nolfo, Tapan Nayak, Alexander Andreopoulos, Guillaume Garreau, Marcela
  Mendoza, et~al.
\newblock A low power, fully event-based gesture recognition system.
\newblock In {\em Proceedings of the IEEE Conference on Computer Vision and
  Pattern Recognition}, pages 7243--7252, 2017.

\bibitem{baby2017dynamic}
Stefanie~Anna Baby, Bimal Vinod, Chaitanya Chinni, and Kaushik Mitra.
\newblock Dynamic vision sensors for human activity recognition.
\newblock In {\em 2017 4th IAPR Asian Conference on Pattern Recognition
  (ACPR)}, pages 316--321. IEEE, 2017.

\bibitem{bertasius2021TimeSformer}
Gedas Bertasius, Heng Wang, and Lorenzo Torresani.
\newblock Is space-time attention all you need for video understanding?
\newblock In {\em ICML}, volume~2, page~4, 2021.

\bibitem{bi2020graph}
Yin Bi, Aaron Chadha, Alhabib Abbas, Eirina Bourtsoulatze, and Yiannis
  Andreopoulos.
\newblock Graph-based spatio-temporal feature learning for neuromorphic vision
  sensing.
\newblock {\em IEEE Transactions on Image Processing}, 29:9084--9098, 2020.

\bibitem{brandli2014240}
Christian Brandli, Raphael Berner, Minhao Yang, Shih-Chii Liu, and Tobi
  Delbruck.
\newblock A 240$\times$ 180 130 db 3 $\mu$s latency global shutter
  spatiotemporal vision sensor.
\newblock {\em IEEE Journal of Solid-State Circuits}, 49(10):2333--2341, 2014.

\bibitem{caba2015activitynet}
Fabian Caba~Heilbron, Victor Escorcia, Bernard Ghanem, and Juan Carlos~Niebles.
\newblock Activitynet: A large-scale video benchmark for human activity
  understanding.
\newblock In {\em Proceedings of the ieee conference on computer vision and
  pattern recognition}, pages 961--970, 2015.

\bibitem{cannici2020mlstm}
Marco Cannici, Marco Ciccone, Andrea Romanoni, and Matteo Matteucci.
\newblock A differentiable recurrent surface for asynchronous event-based data.
\newblock In {\em European Conference on Computer Vision}, pages 136--152.
  Springer, 2020.

\bibitem{ceolini2020hand}
Enea Ceolini, Charlotte Frenkel, Sumit~Bam Shrestha, Gemma Taverni, Lyes
  Khacef, Melika Payvand, and Elisa Donati.
\newblock Hand-gesture recognition based on emg and event-based camera sensor
  fusion: A benchmark in neuromorphic computing.
\newblock {\em Frontiers in Neuroscience}, 14:637, 2020.

\bibitem{chen2021novel}
Guang Chen, Zhongcong Xu, Zhijun Li, Huajin Tang, Sanqing Qu, Kejia Ren, and
  Alois Knoll.
\newblock A novel illumination-robust hand gesture recognition system with
  event-based neuromorphic vision sensor.
\newblock {\em IEEE Transactions on Automation Science and Engineering},
  18(2):508--520, 2021.

\bibitem{chen2019fast}
Huaijin Chen, Wanjia Liu, Rishab Goel, Rhonald~C Lua, Siddharth Mittal, Yuzhong
  Huang, Ashok Veeraraghavan, and Ankit~B Patel.
\newblock Fast retinomorphic event-driven representations for video gameplay
  and action recognition.
\newblock {\em IEEE Transactions on Computational Imaging}, 6:276--290, 2019.

\bibitem{chen2020dyGCN}
Junming Chen, Jingjing Meng, Xinchao Wang, and Junsong Yuan.
\newblock Dynamic graph cnn for event-camera based gesture recognition.
\newblock In {\em 2020 IEEE International Symposium on Circuits and Systems
  (ISCAS)}, pages 1--5. IEEE, 2020.

\bibitem{chen2019celexV}
Shoushun Chen and Menghan Guo.
\newblock Live demonstration: Celex-v: a 1m pixel multi-mode event-based
  sensor.
\newblock In {\em 2019 IEEE/CVF Conference on Computer Vision and Pattern
  Recognition Workshops (CVPRW)}, pages 1682--1683. IEEE, 2019.

\bibitem{clady2017motion}
Xavier Clady, Jean-Matthieu Maro, S{\'e}bastien Barr{\'e}, and Ryad~B Benosman.
\newblock A motion-based feature for event-based pattern recognition.
\newblock {\em Frontiers in neuroscience}, 10:594, 2017.

\bibitem{deng2021mvfnet}
Yongjian Deng, Hao Chen, and Youfu Li.
\newblock Mvf-net: A multi-view fusion network for event-based object
  classification.
\newblock {\em IEEE Transactions on Circuits and Systems for Video Technology},
  2021.

\bibitem{deng2020amae}
Yongjian Deng, Youfu Li, and Hao Chen.
\newblock Amae: Adaptive motion-agnostic encoder for event-based object
  classification.
\newblock {\em IEEE Robotics and Automation Letters}, 5(3):4596--4603, 2020.

\bibitem{dosovitskiy2020ViT}
Alexey Dosovitskiy, Lucas Beyer, Alexander Kolesnikov, Dirk Weissenborn,
  Xiaohua Zhai, Thomas Unterthiner, Mostafa Dehghani, Matthias Minderer, Georg
  Heigold, Sylvain Gelly, et~al.
\newblock An image is worth 16x16 words: Transformers for image recognition at
  scale.
\newblock In {\em International Conference on Learning Representations}, 2020.

\bibitem{fang2021snnresnet}
Wei Fang, Zhaofei Yu, Yanqi Chen, Tiejun Huang, Timoth{\'e}e Masquelier, and
  Yonghong Tian.
\newblock Deep residual learning in spiking neural networks.
\newblock {\em NeurIPS}, 2021.

\bibitem{feichtenhofer2020x3d}
Christoph Feichtenhofer.
\newblock X3d: Expanding architectures for efficient video recognition.
\newblock In {\em Proceedings of the IEEE/CVF Conference on Computer Vision and
  Pattern Recognition}, pages 203--213, 2020.

\bibitem{feichtenhofer2019slowfast}
Christoph Feichtenhofer, Haoqi Fan, Jitendra Malik, and Kaiming He.
\newblock Slowfast networks for video recognition.
\newblock In {\em Proceedings of the IEEE/CVF international conference on
  computer vision}, pages 6202--6211, 2019.

\bibitem{gallegoevent}
Guillermo Gallego, Tobi Delbruck, Garrick~Michael Orchard, Chiara Bartolozzi,
  Brian Taba, Andrea Censi, Stefan Leutenegger, Andrew Davison, Jorg Conradt,
  Kostas Daniilidis, et~al.
\newblock Event-based vision: A survey.
\newblock {\em IEEE transactions on pattern analysis and machine intelligence}.

\bibitem{gehrig2019EST}
Daniel Gehrig, Antonio Loquercio, Konstantinos~G Derpanis, and Davide
  Scaramuzza.
\newblock End-to-end learning of representations for asynchronous event-based
  data.
\newblock In {\em Proceedings of the IEEE/CVF International Conference on
  Computer Vision}, pages 5633--5643, 2019.

\bibitem{george2020reservoir}
Arun~M George, Dighanchal Banerjee, Sounak Dey, Arijit Mukherjee, and P
  Balamurali.
\newblock A reservoir-based convolutional spiking neural network for gesture
  recognition from dvs input.
\newblock In {\em 2020 International Joint Conference on Neural Networks
  (IJCNN)}, pages 1--9. IEEE, 2020.

\bibitem{gu2018ava}
Chunhui Gu, Chen Sun, David~A Ross, Carl Vondrick, Caroline Pantofaru, Yeqing
  Li, Sudheendra Vijayanarasimhan, George Toderici, Susanna Ricco, Rahul
  Sukthankar, et~al.
\newblock Ava: A video dataset of spatio-temporally localized atomic visual
  actions.
\newblock In {\em Proceedings of the IEEE Conference on Computer Vision and
  Pattern Recognition}, pages 6047--6056, 2018.

\bibitem{he2016resnet}
Kaiming He, Xiangyu Zhang, Shaoqing Ren, and Jian Sun.
\newblock Deep residual learning for image recognition.
\newblock In {\em Proceedings of the IEEE conference on computer vision and
  pattern recognition}, pages 770--778, 2016.

\bibitem{kay2017kinetics}
Will Kay, Joao Carreira, Karen Simonyan, Brian Zhang, Chloe Hillier, Sudheendra
  Vijayanarasimhan, Fabio Viola, Tim Green, Trevor Back, Paul Natsev, et~al.
\newblock The kinetics human action video dataset.
\newblock {\em arXiv preprint arXiv:1705.06950}, 2017.

\bibitem{kim2021nimagenet}
Junho Kim, Jaehyeok Bae, Gangin Park, Dongsu Zhang, and Young~Min Kim.
\newblock N-imagenet: Towards robust, fine-grained object recognition with
  event cameras.
\newblock In {\em Proceedings of the IEEE/CVF International Conference on
  Computer Vision}, pages 2146--2156, 2021.

\bibitem{kliper2011action}
Orit Kliper-Gross, Tal Hassner, and Lior Wolf.
\newblock The action similarity labeling challenge.
\newblock {\em IEEE Transactions on Pattern Analysis and Machine Intelligence},
  34(3):615--621, 2011.

\bibitem{kong2018humanARSurvey}
Yu Kong and Yun Fu.
\newblock Human action recognition and prediction: A survey.
\newblock {\em arXiv preprint arXiv:1806.11230}, 2018.

\bibitem{kuehne2011hmdb}
Hildegard Kuehne, Hueihan Jhuang, Est{\'\i}baliz Garrote, Tomaso Poggio, and
  Thomas Serre.
\newblock Hmdb: a large video database for human motion recognition.
\newblock In {\em 2011 International conference on computer vision}, pages
  2556--2563. IEEE, 2011.

\bibitem{lagorce2016hots}
Xavier Lagorce, Garrick Orchard, Francesco Galluppi, Bertram~E Shi, and Ryad~B
  Benosman.
\newblock Hots: a hierarchy of event-based time-surfaces for pattern
  recognition.
\newblock {\em IEEE transactions on pattern analysis and machine intelligence},
  39(7):1346--1359, 2016.

\bibitem{li2018deepCNN}
Hongmin Li, Guoqi Li, Xiangyang Ji, and Luping Shi.
\newblock Deep representation via convolutional neural network for
  classification of spatiotemporal event streams.
\newblock {\em Neurocomputing}, 299:1--9, 2018.

\bibitem{li2017cifar10}
Hongmin Li, Hanchao Liu, Xiangyang Ji, Guoqi Li, and Luping Shi.
\newblock Cifar10-dvs: an event-stream dataset for object classification.
\newblock {\em Frontiers in neuroscience}, 11:309, 2017.

\bibitem{Li2022vidardvsDet}
Jianing Li, Xiao Wang, Lin Zhu, Jia Li, Tiejun Huang, and Yonghong Tian.
\newblock Retinomorphic object detection in asynchronous visual streams.
\newblock In {\em Thirty-Sixth {AAAI} Conference on Artificial Intelligence,
  AAAI 2022, February 22-March 1, 2022}, pages 1332--1340. {AAAI} Press, 2022.

\bibitem{lin2019tsm}
Ji Lin, Chuang Gan, and Song Han.
\newblock Tsm: Temporal shift module for efficient video understanding.
\newblock In {\em Proceedings of the IEEE International Conference on Computer
  Vision}, 2019.

\bibitem{lin2021esimagenet}
Yihan Lin, Wei Ding, Shaohua Qiang, Lei Deng, and Guoqi Li.
\newblock Es-imagenet: A million event-stream classification dataset for
  spiking neural networks.
\newblock {\em Frontiers in neuroscience}, page 1546, 2021.

\bibitem{liu2020unsupervised}
Qianhui Liu, Gang Pan, Haibo Ruan, Dong Xing, Qi Xu, and Huajin Tang.
\newblock Unsupervised aer object recognition based on multiscale
  spatio-temporal features and spiking neurons.
\newblock {\em IEEE transactions on neural networks and learning systems},
  31(12):5300--5311, 2020.

\bibitem{LiuXTM021ijcai}
Qianhui Liu, Dong Xing, Huajin Tang, De Ma, and Gang Pan.
\newblock Event-based action recognition using motion information and spiking
  neural networks.
\newblock In {\em IJCAI 2021, Virtual Event / Montreal, Canada, 19-27 August
  2021}, pages 1743--1749. ijcai.org, 2021.

\bibitem{liu2021videoSwin}
Ze Liu, Jia Ning, Yue Cao, Yixuan Wei, Zheng Zhang, Stephen Lin, and Han Hu.
\newblock Video swin transformer.
\newblock {\em arXiv preprint arXiv:2106.13230}, 2021.

\bibitem{liu2021tam}
Zhaoyang Liu, Limin Wang, Wayne Wu, Chen Qian, and Tong Lu.
\newblock Tam: Temporal adaptive module for video recognition.
\newblock In {\em Proceedings of the IEEE/CVF International Conference on
  Computer Vision}, pages 13708--13718, 2021.

\bibitem{mehr2019action}
Aref~Moqadam Mehr, Saeed~Reza Kheradpisheh, and Hadi Farahani.
\newblock Action recognition using supervised spiking neural networks.
\newblock {\em arXiv preprint arXiv:1911.03630}, 2019.

\bibitem{miao2019neuromorphic}
Shu Miao, Guang Chen, Xiangyu Ning, Yang Zi, Kejia Ren, Zhenshan Bing, and
  Alois Knoll.
\newblock Neuromorphic vision datasets for pedestrian detection, action
  recognition, and fall detection.
\newblock {\em Frontiers in neurorobotics}, 13:38, 2019.

\bibitem{monfort2019moments}
Mathew Monfort, Alex Andonian, Bolei Zhou, Kandan Ramakrishnan, Sarah~Adel
  Bargal, Tom Yan, Lisa Brown, Quanfu Fan, Dan Gutfreund, Carl Vondrick, et~al.
\newblock Moments in time dataset: one million videos for event understanding.
\newblock {\em IEEE transactions on pattern analysis and machine intelligence},
  42(2):502--508, 2019.

\bibitem{orchard2015converting}
Garrick Orchard, Ajinkya Jayawant, Gregory~K Cohen, and Nitish Thakor.
\newblock Converting static image datasets to spiking neuromorphic datasets
  using saccades.
\newblock {\em Frontiers in neuroscience}, 9:437, 2015.

\bibitem{panda2018learning}
Priyadarshini Panda and Narayan Srinivasa.
\newblock Learning to recognize actions from limited training examples using a
  recurrent spiking neural model.
\newblock {\em Frontiers in neuroscience}, 12:126, 2018.

\bibitem{paszke2019pytorch}
Adam Paszke, Sam Gross, Francisco Massa, Adam Lerer, James Bradbury, Gregory
  Chanan, Trevor Killeen, Zeming Lin, Natalia Gimelshein, Luca Antiga, et~al.
\newblock Pytorch: An imperative style, high-performance deep learning library.
\newblock {\em Advances in neural information processing systems}, 32, 2019.

\bibitem{posch2010qvga}
Christoph Posch, Daniel Matolin, and Rainer Wohlgenannt.
\newblock A qvga 143 db dynamic range frame-free pwm image sensor with lossless
  pixel-level video compression and time-domain cds.
\newblock {\em IEEE Journal of Solid-State Circuits}, 46(1):259--275, 2010.

\bibitem{samadzadeh2020convsnn}
Ali Samadzadeh, Fatemeh Sadat~Tabatabaei Far, Ali Javadi, Ahmad Nickabadi, and
  Morteza~Haghir Chehreghani.
\newblock Convolutional spiking neural networks for spatio-temporal feature
  extraction.
\newblock {\em arXiv preprint arXiv:2003.12346}, 2020.

\bibitem{serrano2015poker}
Teresa Serrano-Gotarredona and Bernab{\'e} Linares-Barranco.
\newblock Poker-dvs and mnist-dvs. their history, how they were made, and other
  details.
\newblock {\em Frontiers in neuroscience}, 9:481, 2015.

\bibitem{sigurdsson2016hollywood}
Gunnar~A Sigurdsson, G{\"u}l Varol, Xiaolong Wang, Ali Farhadi, Ivan Laptev,
  and Abhinav Gupta.
\newblock Hollywood in homes: Crowdsourcing data collection for activity
  understanding.
\newblock In {\em European Conference on Computer Vision}, pages 510--526.
  Springer, 2016.

\bibitem{sironi2018hats}
Amos Sironi, Manuele Brambilla, Nicolas Bourdis, Xavier Lagorce, and Ryad
  Benosman.
\newblock Hats: Histograms of averaged time surfaces for robust event-based
  object classification.
\newblock In {\em Proceedings of the IEEE Conference on Computer Vision and
  Pattern Recognition}, pages 1731--1740, 2018.

\bibitem{song2019TSM}
Xiaolin Song, Cuiling Lan, Wenjun Zeng, Junliang Xing, Xiaoyan Sun, and Jingyu
  Yang.
\newblock Temporal--spatial mapping for action recognition.
\newblock {\em IEEE Transactions on Circuits and Systems for Video Technology},
  30(3):748--759, 2019.

\bibitem{soomro2012ucf101}
Khurram Soomro, Amir~Roshan Zamir, and Mubarak Shah.
\newblock Ucf101: A dataset of 101 human actions classes from videos in the
  wild.
\newblock {\em arXiv preprint arXiv:1212.0402}, 2012.

\bibitem{sutskever2013SGD}
Ilya Sutskever, James Martens, George Dahl, and Geoffrey Hinton.
\newblock On the importance of initialization and momentum in deep learning.
\newblock In {\em International conference on machine learning}, pages
  1139--1147. PMLR, 2013.

\bibitem{tran2015c3d}
Du Tran, Lubomir Bourdev, Rob Fergus, Lorenzo Torresani, and Manohar Paluri.
\newblock Learning spatiotemporal features with 3d convolutional networks.
\newblock In {\em Proceedings of the IEEE international conference on computer
  vision}, pages 4489--4497, 2015.

\bibitem{tran2018R2Plus1D}
Du Tran, Heng Wang, Lorenzo Torresani, Jamie Ray, Yann LeCun, and Manohar
  Paluri.
\newblock A closer look at spatiotemporal convolutions for action recognition.
\newblock In {\em Proceedings of the IEEE conference on Computer Vision and
  Pattern Recognition}, pages 6450--6459, 2018.

\bibitem{wang2019steventclouds}
Qinyi Wang, Yexin Zhang, Junsong Yuan, and Yilong Lu.
\newblock Space-time event clouds for gesture recognition: From rgb cameras to
  event cameras.
\newblock In {\em 2019 IEEE Winter Conference on Applications of Computer
  Vision (WACV)}, pages 1826--1835. IEEE, 2019.

\bibitem{wang2021visevent}
Xiao Wang, Jianing Li, Lin Zhu, Zhipeng Zhang, Zhe Chen, Xin Li, Yaowei Wang,
  Yonghong Tian, and Feng Wu.
\newblock Visevent: Reliable object tracking via collaboration of frame and
  event flows.
\newblock {\em arXiv preprint arXiv:2108.05015}, 2021.

\bibitem{wang2019evGait}
Yanxiang Wang, Bowen Du, Yiran Shen, Kai Wu, Guangrong Zhao, Jianguo Sun, and
  Hongkai Wen.
\newblock Ev-gait: Event-based robust gait recognition using dynamic vision
  sensors.
\newblock In {\em Proceedings of the IEEE/CVF Conference on Computer Vision and
  Pattern Recognition}, pages 6358--6367, 2019.

\bibitem{wang2021eventGNN}
Yanxiang Wang, Xian Zhang, Yiran Shen, Bowen Du, Guangrong Zhao, Lizhen Cui~Cui
  Lizhen, and Hongkai Wen.
\newblock Event-stream representation for human gaits identification using deep
  neural networks.
\newblock {\em IEEE Transactions on Pattern Analysis and Machine Intelligence},
  2021.

\bibitem{wang2021actionnet}
Zhengwei Wang, Qi She, and Aljosa Smolic.
\newblock Action-net: Multipath excitation for action recognition.
\newblock In {\em Proceedings of the IEEE/CVF Conference on Computer Vision and
  Pattern Recognition}, pages 13214--13223, 2021.

\bibitem{wu2020multipath}
Xiao Wu and Junsong Yuan.
\newblock Multipath event-based network for low-power human action recognition.
\newblock In {\em 2020 IEEE 6th World Forum on Internet of Things (WF-IoT)},
  pages 1--5. IEEE, 2020.

\bibitem{xie2022vmvgcn}
Bochen Xie, Yongjian Deng, Zhanpeng Shao, Hai Liu, and Youfu Li.
\newblock Vmv-gcn: Volumetric multi-view based graph cnn for event stream
  classification.
\newblock {\em IEEE Robotics and Automation Letters}, 7(2):1976--1983, 2022.

\bibitem{xing2020new}
Yannan Xing, Gaetano Di~Caterina, and John Soraghan.
\newblock A new spiking convolutional recurrent neural network (scrnn) with
  applications to event-based hand gesture recognition.
\newblock {\em Frontiers in Neuroscience}, 14:1143, 2020.

\bibitem{yao2021TASNN}
Man Yao, Huanhuan Gao, Guangshe Zhao, Dingheng Wang, Yihan Lin, Zhaoxu Yang,
  and Guoqi Li.
\newblock Temporal-wise attention spiking neural networks for event streams
  classification.
\newblock In {\em Proceedings of the IEEE/CVF International Conference on
  Computer Vision}, pages 10221--10230, 2021.

\bibitem{zhu2018evflownet}
Alex~Zihao Zhu and Liangzhe Yuan.
\newblock Ev-flownet: Self-supervised optical flow estimation for event-based
  cameras.
\newblock In {\em Robotics: Science and Systems}, 2018.

\bibitem{zhu2021neuspike}
Lin Zhu, Jianing Li, Xiao Wang, Tiejun Huang, and Yonghong Tian.
\newblock Neuspike-net: High speed video reconstruction via bio-inspired
  neuromorphic cameras.
\newblock In {\em Proceedings of the IEEE/CVF International Conference on
  Computer Vision}, pages 2400--2409, 2021.

\bibitem{zhu2022eventsnn}
Lin Zhu, Xiao Wang, Yi Chang, Jianing Li, Tiejun Huang, and Yonghong Tian.
\newblock Event-based video reconstruction via potential-assisted spiking
  neural network.
\newblock In {\em Proceedings of the IEEE/CVF Conference on Computer Vision and
  Pattern Recognition}, pages 3594--3604, 2022.

\end{thebibliography}
}

\end{document}